\documentclass[]{fairmeta}
\usepackage{float}
\usepackage[table]{xcolor}
\usepackage{colortbl}

\newcommand{\methodname}{\texttt{MobileLLM-Pro}}

\title{MobileLLM-Pro Technical Report}

\author[*]{Patrick Huber}
\author[*]{Ernie Chang}
\author[*]{Wei Wen}
\author[*]{Igor Fedorov}
\author[]{Tarek Elgamal}
\author[]{Hanxian Huang}
\author[]{Naveen Suda}
\author[]{Chinnadhurai Sankar}
\author[]{Vish Vogeti}
\author[]{Yanghan Wang}
\author[]{Alex Gladkov}
\author[]{Kai Sheng Tai}
\author[]{Abdelrahman Elogeel}
\author[]{Tarek Hefny}
\author[]{Vikas Chandra}
\author[]{Ahmed Aly}
\author[]{Anuj Kumar}
\author[\dagger]{Raghuraman Krishnamoorthi}
\author[\dagger*]{Adithya Sagar}

\affiliation[]{Meta Reality Labs}

\contribution[*]{Joint First Author}
\contribution[\dagger]{Joint last author}

\abstract{
Efficient on-device language models around 1 billion parameters are essential for powering low-latency AI applications on mobile and wearable devices. However, achieving strong performance in this model class, while supporting long context windows and practical deployment remains a significant challenge. We introduce \methodname, a 1-billion-parameter language model optimized for on-device deployment. \methodname~achieves state-of-the-art results across 11 standard benchmarks, significantly outperforming both Gemma 3-1B and Llama 3.2-1B, while supporting context windows of up to 128,000 tokens and showing only minor performance regressions at 4-bit quantization.
These improvements are enabled by four core innovations: (1) implicit positional distillation, a novel technique that effectively instills long-context capabilities through knowledge distillation; (2) a specialist model merging framework that fuses multiple domain experts into a compact model without parameter growth; (3) simulation-driven data mixing using utility estimation; and (4) 4-bit quantization-aware training with self-distillation. We release our model weights and code to support future research in efficient on-device language models.}

\date{\today}
\correspondence{\email{patrickhuber@meta.com}, \email{adithyasagar@meta.com}}
\metadata[Code]{\url{https://huggingface.co/collections/facebook/mobilellm-pro}}

\begin{document}

\maketitle

\section{Introduction}
\label{section:intro}
Foundational large language models (LLMs) continue to push the boundaries of AI capabilities through increasingly large-scale architectures, achieving state-of-the-art results across diverse tasks at the cost of substantial computational resources and inference latency \citep{brown2020language, touvron2024llama3,gemmateam2025gemma3technicalreport}. In parallel, a growing body of research has emerged focused on compact, efficient models in the sub-2B parameter range that aim to preserve strong generalist capabilities while dramatically reducing memory footprint and computational demands.  \citep{gemmateam2025gemma3technicalreport, touvron2024llama3, yang2025qwen3technicalreport, allal2025smollm2smolgoesbig}. Beyond cost efficiency, these compact models enable practical on-device applications: improved latency by eliminating server round-trips, offline functionality without network connectivity, and seamless integration on resource-constrained devices including smartphones, laptops, and wearables.

 Critically, the synergy between powerful server-side models and efficient on-device models is essential for realizing truly ubiquitous, always-present AI systems that can dynamically adapt to connectivity, latency, and privacy requirements.

\begin{figure}
    \centering
    \includegraphics[width=1.0\linewidth]{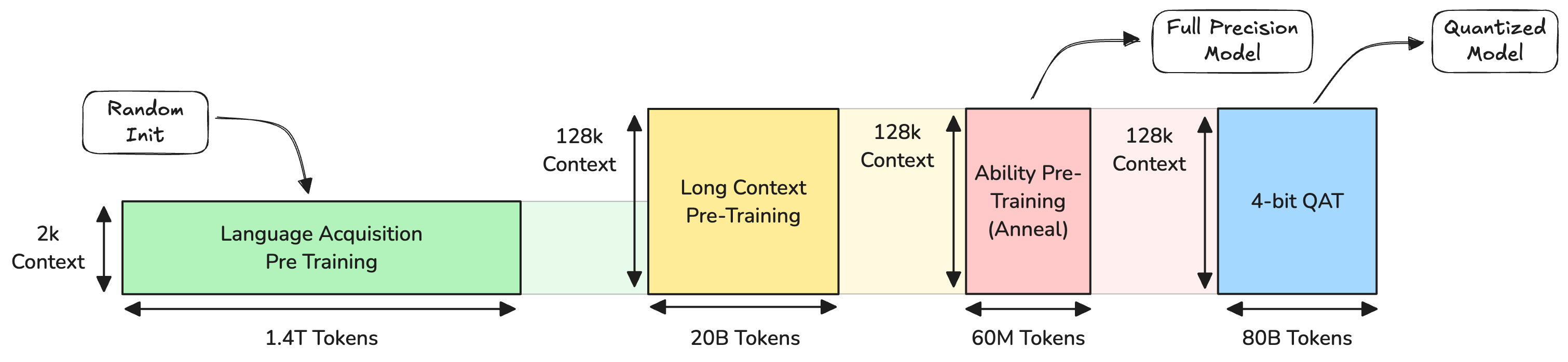}
    \caption{Schematic of our four-staged pre-training approach with three performance-specific pre-training phases and an additional quantization stage.}
    \label{fig:pt_stages}
\end{figure}

However, realizing the full potential of on-device foundational models requires addressing several critical challenges. First, small-scale models tend to be less robust during training compared to their larger counterparts, making them more susceptible to performance degradation due to data quality issues and distribution shifts across training phases. Second, extending long-context capabilities beyond 32k, which is common in frontier models, remains relatively unexplored \citep{gemmateam2025gemma3technicalreport, yang2025qwen3technicalreport, allal2025smollm2smolgoesbig}. Third, practical deployment demands efficient compression of both, the model weights and the runtime memory footprint, without sacrificing performance. \citep{frantar2022gptq, fedorov2024llamaguard31bint4compact}.

In this work, we introduce \methodname, a 1 billion parameter foundational language model achieving state-of-the-art (SOTA) pre-training performance at its size as well as competitive instruction-tuning results. In pre-training, \methodname~surpasses Gemma 3-1B \citep{gemmateam2025gemma3technicalreport} and Llama 3.2-1B \citep{touvron2024llama3} on eleven widely-used knowledge and reasoning benchmarks. We achieve this by combining established techniques, such as local-global attention, knowledge distillation and phased training, with novel innovations tailored for on-device deployment. Our contributions  are the following:
\begin{enumerate}
\item \textbf{Simulation-driven data mixing}: We employ an Automixer style data mixing framework in all phases to address the data sensitivity of compact models by leveraging simulation runs to guide data composition  \cite{chang2025automixercheckpointartifactsautomatic}.
\item \textbf{Implicit positional distillation}: To mitigate data distribution shifts between the initial pre-training phase and the long-context extension phase, we introduce implicit positional distillation, where a large teacher model transfers long context abilities without exposing the student to long context data.
\item \textbf{Specialist model merging}: We introduce a new model averaging technique,  where we train multiple specialist checkpoints in parallel during pretraining, each optimized for different capability (such as coding, reasoning).  We then use non-uniform weight averaging to combine these specialists into a single model, improving the overall performance without any parameter overhead.
\item \textbf{Quantization-aware training}: We quantize the model to support CPU and accelerator (ANE/HTP) inference. Using quantization range learning and self-distillation with the floating point teacher, we maintain long-context and reasoning capabilities at the 4-bit scale.

We additionally instruction tune \methodname to build an on-device assistant style model capable of supporting common mobile centric interactions such as summarization, rewriting, simple function calling and question-answering

To support future research we release all model checkpoints at \url{https://huggingface.co/collections/facebook/mobilellm-pro}
\end{enumerate}

\begin{table}[t]
\centering
\begin{tabular}{@{}ll@{}}
\toprule
\textbf{Architecture} &  \\ \midrule
Layers & 30 \\
Attention Heads & 20 \\
KV Heads & 4 \\
Dimension & 1280 \\
Hidden Dimension & 6144 \\
Vocabulary Size & 202,048 \\
Total Parameters & 1,084M (1.08B) \\ \midrule
\textbf{Modality} &  \\ \midrule
Input Modality & Text \\
Output Modality & Text \\
Languages & English \\ \midrule
\textbf{Other} &  \\ \midrule
Context Length & 128k tokens \\
Embeddings & Shared \\
Attention & Local-Global: 512 local attention, \\
          & global attention every 4th layer \\ 
\bottomrule
\end{tabular}
\caption{Summary of the \methodname~Model Specifications}
\label{tab:specs}
\end{table}

\section{Model Architecture}
\label{sec:arch}
The \methodname~model architecture is based on the standard transformer framework \citep{vaswani2017attention}, incorporating several design elements from the Llama series \citep{touvron2023llama, touvron2024llama3}.  Despite containing only 1B parameters, \methodname~features the full 202,048 vocabulary used in frontier models, such as Llama4. To reduce the substantial parameter allocation required for the input embedding and language modeling head, we employ embedding sharing \citep{press2017using, liu2024mobilellmoptimizingsubbillionparameter}, reducing the embedding footprint by 260 million parameter ($\approx$25\% of the overall model). \methodname~consists of 30 transformer layers, balancing performance with latency \citep{liu2024mobilellmoptimizingsubbillionparameter}. With a hidden dimension of 1280, each transformer layer in our model contains 20 distinct attention heads of dimension 64. We further use Grouped-Query-Attention (GQA) \citep{ainslie-etal-2023-gqa} with 4 key-value (kv) pairs. The feed-forward network (FFN) uses 4.8x up-scaling to enhance model capacity and expressiveness. Importantly, while most on-device models limit their context length to less than 8,000, \citep{allal2025smollm2smolgoesbig} or 32,000 \citep{yang2025qwen3technicalreport,gemmateam2025gemma3technicalreport} tokens, we support a context length of up to 128,000 tokens. This enables \methodname~to process sequences four times longer than Gemma 3-1B \citep{gemmateam2025gemma3technicalreport} and Qwen 3-0.6B \citep{yang2025qwen3technicalreport}, matching the context window of Llama 3.2-1B \citep{touvron2024llama3}. With the extended context length, the size of the model increases linearly, easily exceeding hardware constraints and making the model deployment on edge devices challenging. To address this, we interleave local and global attention across layers of our model in a 3 to 1 ratio (i.e. one global attention layer for every 3 local layers) while ensuring the first and final layers to be global layers. We limit the attention computation in local attention layers to the sliding window of the most recent 512 tokens, balancing long-context abilities with computational efficiency \citep{beltagy2020longformer, zaheer2020bigbird, gemmateam2025gemma3technicalreport}. For more details on the model architecture, see Table \ref{tab:specs}.

\section{Phases of Pre-training}
\label{section:pre_train}

The pre-training phase of foundational text models is critical for developing core language capabilities, a key driver for understanding and generation \citep{brown2020language}. 
As a result, pre-training has been extensively studied for large-scale foundational models. 
For smaller model variants, especially those below 3B parameters, the exploration of pre-training dynamics has been significantly less thorough, which can lead to sub-optimal setups. 
Specifically studying on-device sized foundational language models, we propose a four-phased pre-training strategy informed by extensive ablations at on-device sizes, as outlined in Figure \ref{fig:pt_stages}.
\begin{itemize}
    \item \textbf{Phase 1}: Language-acquisition focusing on general language learning. This phase is especially sensitive to the training data-mix. Containing the largest token budget throughout pre-training, we primarily focus on a well balanced data composition, as described in section \ref{sec:ph1}
    \item \textbf{Phase 2}: Context-Expansion to enable long context abilities. In this phase, we extend the pre-trained checkpoint to support long context queries (128,000 tokens) using ``implicit positional distillation''. For more information, see section \ref{sec:ph2}    
    \item \textbf{Phase 3}: Specialist Model Merging. In the third phase, we anneal the model by training multiple parallel specialists, subsequently merged back into a single checkpoint (see section \ref{sec:ph3})
    \item \textbf{Phase 4}: Towards Efficiency. In the final pre-training stage, we use Quantization Aware Training (QAT) to reduce the model's memory footprint, making them true on-device candidates. For more details, see section \ref{sec:ph4}
\end{itemize}
Throughout all pre-training phases, we employ logit-based knowledge distillation (KD) \citep{hinton2015distilling, jiao2020tinybert} from a large teacher model (Llama 4-Scout). Compared to standard pre-training, usually computing the cross-entropy (CE) loss between samples and predictions, we utilize the forward KL-divergence loss between the teacher and student logits. This way, instead of a one-hot target from the vocabulary, the models learns from the much richer logit signal, guiding the student model towards the better language modeling target.

\section{Phase 1: Language-Acquisition via Offline Data-Mix Simulations}
\label{sec:ph1}
The language-acquisition stage is the first and longest phase of model training (1.4T tokens). 
Starting from randomly initialized weights, the model learns the foundational structure of its weight space, which has a lasting impact on all subsequent training stages. 
Table~\ref{tab:pt_mix} shows the Stage 1 and 2 pre-training data-mix. 
Following the Scalable Data Mixer (SDM)~\cite{chang2025sdm}, we employ a dynamic data-mixing strategy that automatically estimates the relative contribution of different data domains through a single simulated training run. 
The considered domains include general language data, knowledge, reasoning, code, and math. 

To arrive at the final pre-training data-mix, we model the influence of each data sample as a function of its domain-specific information gain and predicted downstream utility. During the simulation, lightweight statistical language models (SLM) are maintained for each domain to quantify the influence of a sample $x$. Specifically, the influence of sample $x$ on a domain $\mathcal{D}$ is measured by its effect on the cross-entropy loss across samples within $\mathcal{D}$ (i.e. by calculating the loss difference of samples in $\mathcal{D}$ before and after updating the SLM with $x$). 
Individual, domain-level influence scores are then aggregated into an influence vector $\Delta$ combining all domain-specific influence scores.
The influence vector $\Delta$ is then passed through a lightweight neural regressor to estimate the downstream utility $\hat{y}$ of each domain. 
The resulting utility estimates provide a principled measure of the expected improvement when prioritizing samples from a given data source. 
These estimates can then be used to construct balanced and capability-aware data mixtures for large-scale pre-training\footnote{For more details refer to~\citet{chang2025sdm}.}.

\begin{table}[h!]
\centering
\begin{tabular}{l|rrr}
\textbf{Dataset} & \textbf{Rows (M)} & \textbf{Tokens (B)} & \textbf{Weight (\%)} \\
\hline
Fineweb-EDU \citeyearpar{penedo2024fineweb} & 1279.1 & 1300 & 89.75 \\
Starcoder \citeyearpar{li2023starcoder} & 206.6 & 263.8 & 4.66 \\
Open Web Math \citeyearpar{paster2023openwebmath} & 6.1 & 12.6 & 1.92 \\
Arxiv \citeyearpar{together2023redpajama} & 1.5 & 28 & 1.35 \\
Wiki & 7.2 & 3.7 & 1.02 \\
Stack Exchange \citeyearpar{together2023redpajama} & 29.2 & 19.6 & 1.02 \\
Algebraic Stack \citeyearpar{azerbayev2023llemma} & 3.4 & 12.6 & 0.24 \\
\hline
\textbf{Total} & \textbf{1500} & \textbf{1640.3} & \textbf{100} \\
\end{tabular}
\caption{Dataset statistics used in our Stage 1 and Stage 2 pre-training data-mix.}
\label{tab:pt_mix}
\end{table}

To efficiently determine the optimal composition of training data, we perform a single offline simulation in which the model processes the entire corpus once, estimating the predicted utility $\hat{y}$ of each data source $\mathcal{D}$. 
Rather than adjusting sampling probabilities dynamically during training, these utility values are used to derive static sampling weights that approximate the steady-state behavior of an adaptive system. 
The resulting weights $w_i$ define the relative sampling frequency of each domain in the final mixture (see left-most column in Table \ref{tab:pt_mix}). 
This formulation effectively captures the adaptive dynamics by prioritizing high-utility domains while maintaining sufficient domain diversity. 
By constructing the mixture offline, the method avoids the computational cost of online feedback during training, offering an efficient approach to scalable, capability-aware pre-training.
Table \ref{tab:pt_mix} shows the resulting data-mix used for \methodname~estimated with SDM after 1 million steps. 

Based on these insights, we train our 1B parameter model defined in section \ref{sec:arch} on 1.4T tokens using the data distribution described in Table \ref{tab:pt_mix}. Given our effective batch-size of 2M tokens, we pre-train our model using $640,000$ weight update steps. We set the max learning rate as $4e-4$ with $10,000$ warm-up steps and a final learning rate of $0.0$ using cosine learning rate scheduling. For the knowledge distillation setup, we use the complete $202,048$ element logit from the Llama 4-Scout teacher model as the training signal.

\section{Phase 2: Context-Expansion using Implicit Positional Distillation}
\label{sec:ph2}
\begin{figure}[h!]
    \centering
    \includegraphics[width=1.0\linewidth]{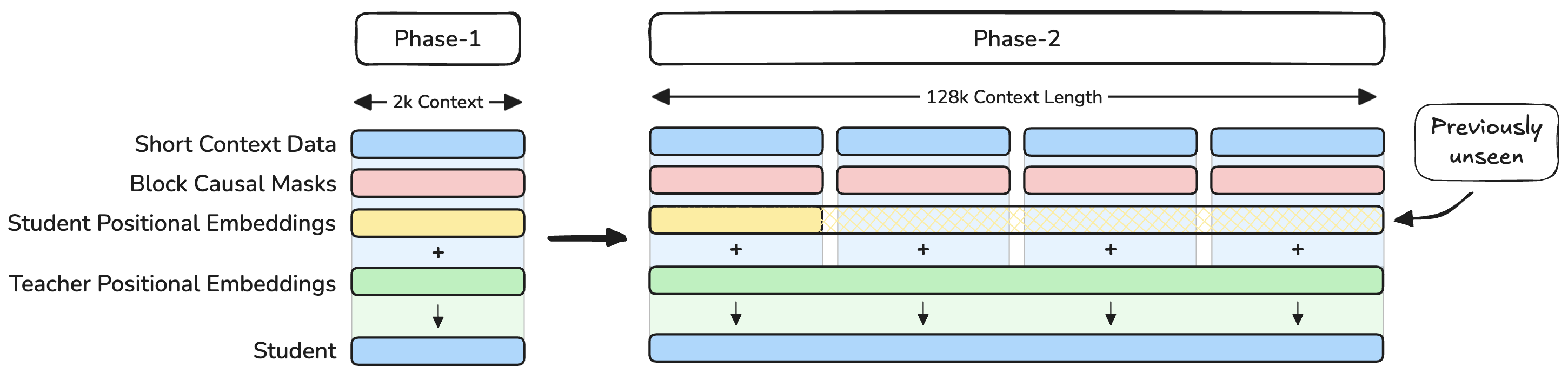}
    \caption{Schematic of the first two pre-training phases, showing the lack of long context positional embeddings in the student model at the beginning of Phase 2. Training the student using teacher logits, which implicitly contain long-context positional information, we are able to teach the student model to learn long-context relationships}
    \label{fig:long_short_cxt}
\end{figure}

Enabling on-device models for long-context inference has become increasingly important for real-world applications such as in-context information retrieval, summarization, and personalization. However, training small models with long-context abilities presents significant challenges during training and inference. During inference, memory requirements scale with context length, creating bottlenecks for on-device deployment. We address this using local-global attention mechanisms that reduce per-token memory overhead \citep{beltagy2020longformer}. During long-context training, the main bottleneck becomes how to effectively extend a model's context window while preserving previously learned information. This problem often manifests due to the large distribution shift when transitioning from general language modeling (Phase 1) to long-context specialization (Phase 2), where the model must simultaneously learn new positional relationships while maintaining its existing knowledge (see Figure \ref{fig:long_short_cxt} for a visual schematic on the differences between Phase 1 and Phase 2). 

To overcome this bottleneck, we demonstrate that knowledge distillation (KD) using a large-scale, long-context trained teacher model (here: Llama 4-Scout) enables small student models to inherit long-context capabilities without direct exposure to long-context data. This, in turn, avoids any data distribution shift between pre-training phases, enabling the model to retain its original performance while extending the context length. We believe that this phenomenon, which we term ``implicit positional distillation'', works through the synergistic interaction of Rotary Position Embedding (RoPE), block causal masking, and logit distillation.

The key insight here is that block causal masking, while preventing information leakage between concatenated documents within a packed batch element, does crucially not constrain the transfer of positional information when using logit-based knowledge distillation. Specifically, when using RoPE positional embeddings, the positional information is encoded through rotational transformations applied to query and key hidden vectors defined as:

\begin{equation}
\begin{bmatrix}
x'_{i} \\
x'_{i+1}
\end{bmatrix}
=
\begin{bmatrix}
\cos(\theta_i p) & -\sin(\theta_i p) \\
\sin(\theta_i p) & \cos(\theta_i p)
\end{bmatrix}
\begin{bmatrix}
x_{i} \\
x_{i+1}
\end{bmatrix}
\end{equation}

Where $\theta_i = 10000^{-\frac{2i}{d}}$ are the rotational frequencies for feature pairs $(i, i+1)$ in the $d$-dimensional hidden state, and $p$ is the absolute position.

\begin{figure}
    \centering
    \includegraphics[width=0.8\linewidth]{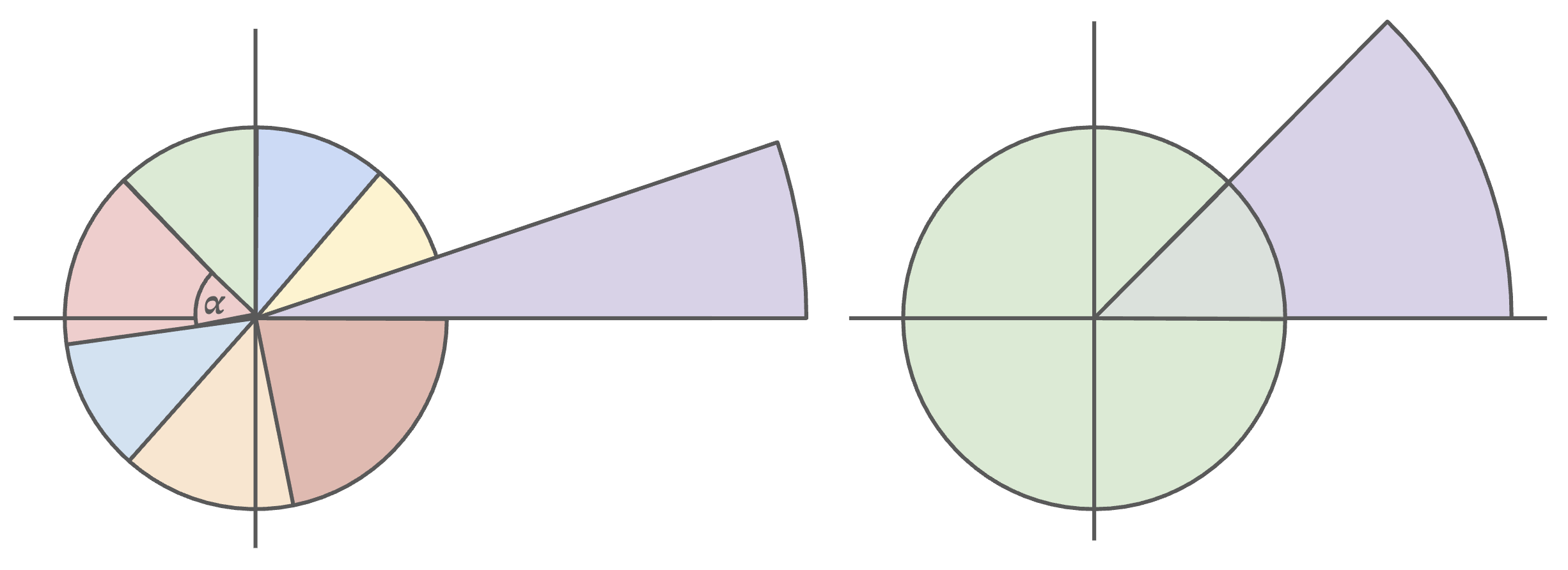}
    \caption{Exploration of the RoPE angular space using concatenated short-context data (left) and true long-context data (right). Purple=Short-context first stage training, Radius=Number of samples trained at the angle in the polar space, $\alpha$=Maximal rotational distance without the teacher model, limited by the short-context training data.}
    \label{fig:ang_space}
\end{figure}

During short-context pre-training, the model only explores a limited portion of the full rotational space. Specifically, with maximum context length $L_{short}$, the model experiences rotations spanning angles from $0$ to $\theta_i \cdot L_{short}$ for each frequency $\theta_i$. This constrains the model's direct experience to a subset of possible positional relationships (Purple in Figure \ref{fig:ang_space}). Crucially, simply extending the context window to cover the full angular space (here: 360 degree = 128,000 tokens) while training on short documents does not solve the long-context problem. Even when concatenating multiple short documents to fill a longer context window, the meaningful semantic and syntactic relationships within each document remain local, i.e. $\theta_{max}-\theta_{min} = \alpha$ (see Figure \ref{fig:ang_space}, left). This means that while RoPE can technically encode larger angular distances, the model never learns meaningful patterns at these distances because no genuine long-range dependencies exist in the training data. As a result, the relationship between any token further apart than the document length (e.g. $\alpha$) provides no useful learning signal for understanding long-range dependencies.

Utilizing ``Implicit Positional Distillation'', true long-context relationships can be learned through positional knowledge transfer from the teacher's logits, without requiring the student to ever directly process long sequences. When the teacher model produces probability distributions over the vocabulary (i.e. logits), this distribution implicitly encodes the positional relationships the teacher has learned during its own training, covering the complete angular RoPE space (see Figure \ref{fig:ang_space}, right). The student model, in turn, learns to mimic these distributions, effectively inheriting the teacher's understanding of how positional distance beyond $\alpha$ affects token relationships.
As a result, we believe that ``implicit positional distillation'' allows student models to acquire long-context capabilities by learning both, token-level predictions and the implicit positional relationships encoded in a teacher's output. This effectively bridges the gap between short-context training and long-context inference requirements, without the need for a data distribution shift.

Due to the ``Implicit Positional Distillation'' and the resulting ability to utilize the original Phase 1 data-mix, the model does not experience any data distribution shift between Phases 1 and 2. As a result, the long-context extension phase can be rather short. Specifically, we train the final Phase 1 checkpoint for another 100,000 steps, exposing the model to 20 billion additional tokens (from the same pre-training data distribution). To account for the larger context length per-sample, we reduce the batch-size and increase model parallelism, resulting in overall smaller effective batch-sizes in this stage. In this second stage, we reduce the max learning rate to $4e-5$ and again use cosine learning rate scheduling to anneal the learning rate to $0.0$.

\section{Phase 3: Specialist Model Merging}
\label{sec:ph3}
\begin{figure}
    \centering
    \includegraphics[width=0.75\linewidth]{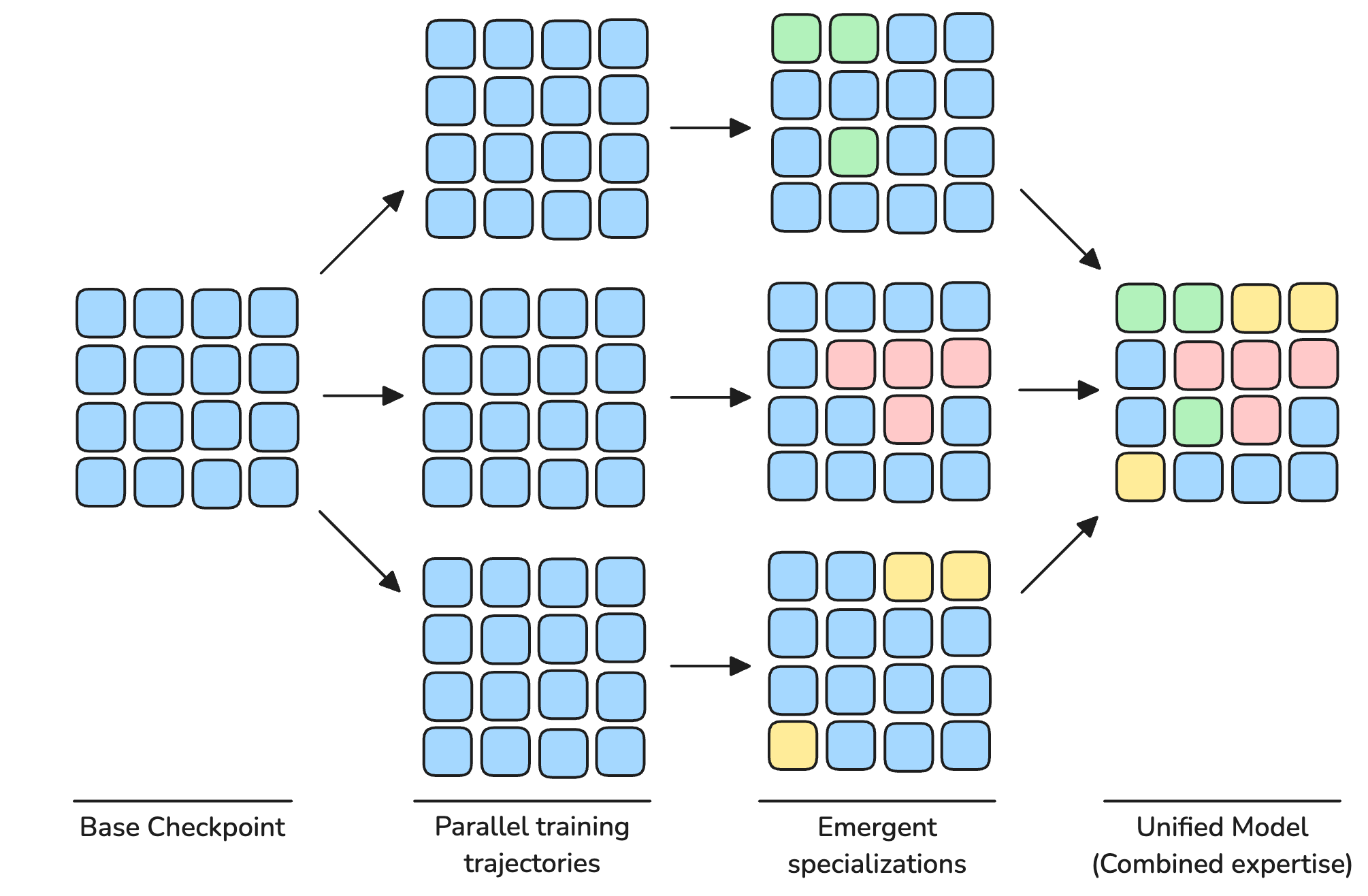}
    \caption{Schematic of the Specialist Model Merging approach, showing the initial base model checkpoint being branched out into parallel training trajectories. Each trajectory is learning emergent specializations in isolation, before they are merged into the final, unified model combining the trajectory expertise.}
    \label{fig:btm}
\end{figure}

The third phase of pre-training focuses on presenting the model with high-quality, domain-centered data samples. Along those lines, we cluster unstructured pre-training data and use our novel ``Specialist Model Merging'' methodology to train multiple, parallel domain experts, similar in spirit to \citet{izmailov2019averagingweightsleadswider} and \citet{li2022branch}. Using this specialist merging paradigm in the final annealing stage, we aim to better combine critical model skills and avoid direct competition between abilities (e.g. reasoning and factuality) during the training process. To this end, we train multiple parallel model instances of the final Phase 2 checkpoint, while gradually reducing the learning rate from $1e-5$ linearly down to zero. Figure \ref{fig:btm} shows a visual representation of the process. We select the optimal model candidates from the ensemble of specialists by \textit{non-uniform weight averaging} to create a single, unified model. The unified model performance, surprisingly, does better than individual specialists in most cases.

Critically, we find that specialist model merging only works if two conditions are met: (1) The model has to already be in a stable weight-space (i.e. this method is only effective in later pre-training stages) and (2) The model parallel updates have to be small, to ensure model weights don't divert too far from the initial model weight distribution. 

Given these two requirements, ``Specialist Model Merging'' perfectly lends itself for the final annealing stage. Here, we train each parallel expert for 60 million tokens distributed over 500 model update steps while keeping the extended context length from Phase 2.

More formally, the Specialist Model Merging paradigm creates $B=(b_1, b_2, ..., b_n)$ parallel specialists for the $n = |B|$ distinct domains $\mathcal{D}_{b=1}^n$, each initialized from the final Stage 2 model state. Each domain specialist $b_i \in B$ is independently trained on a single domain $\mathcal{D}_i$. Once all $n$ specialists are trained, we merge the final checkpoints via \textit{non-uniform weighted parameter averaging}, using per-domain weights $w_b$ to produce a final model $M$:

\begin{equation}
M = \frac{1}{n} \sum_{b=1}^n \theta_b * w_b
\end{equation}

where $\theta_b$ represents the union of all weight matrices of the final checkpoint of branch $b$ and $\sum_{b=1}^n w_b = 1.0$. Overall, the combination of parallel specialists, linear model annealing, and non-uniform weighted parameter averaging produces a robust pre-trained language model that effectively learns the union of all annealing domains (e.g. reasoning, factuality, question-answering) while retaining the generality of the final Phase 2 checkpoint.

We call this checkpoint \methodname\texttt{-base} and make it available at \url{https://huggingface.co/facebook/MobileLLM-Pro-base}

\section{Phase 4: Towards Efficiency -- Quantization Aware Training}
\label{sec:ph4}
Besides the model quality targeted in the pre-training Phases 1-3, we aim to further enable our model to run on both mobile CPUs and accelerators. To enable \methodname~to fit mobile hardware constraints, we employ two quantization schemes for CPU and accelerator support.

\paragraph{CPU -- 4-bit group-wise quantization}On CPU, we quantize weights group-wise to the INT4 data-type. This way, every group of 32 weights shares a common quantization scale. This quantization scheme is best suited to run on CPU backends like \href{https://docs.pytorch.org/executorch/0.5/native-delegates-executorch-xnnpack-delegate.html}{xnnpack}, supporting fine-grained, group-wise quantization. For activations, we apply 8-bit quantization, where the inputs to each linear layer are quantized to INT8 with asymmetric per-token dynamic quantization, described as follows:
    \begin{align}\label{eq:weight quantization}
        Q(x_{in}) &= s_x \times \text{clip} \left(\text{round}\left(\frac{x_{in} - z}{s_x} \right),0,255 \right) + z, \; \; s_x = \frac{1}{255} \left(\max x_{in} - \min x_{in} \right), \; \; z = \min x_{in}
    \end{align}
where $x_{in}$ is an activation vector corresponding to a particular token index. In this dynamic quantization scenario, the tensor is quantized using the per-token min/max range right before the matrix-multiplication operation. We further quantize the KV cache to 8-bits with asymmetric per-token quantization.

\paragraph{Accelerator -- 4-bit channel-wise quantization}Some accelerators such as Apple Neural Engine (\href{https://apple.github.io/coremltools/docs-guides/source/opt-quantization-overview.html}{ANE}) and Qualcomm’s Hexagon Tensor Processor (\href{https://docs.qualcomm.com/bundle/publicresource/topics/80-63442-50/htp_guidelines_int4_weights.html}{HTP}) either do not support group-wise quantization or implement it inefficiently. As a result, we use channel-wise quantization for accelerators, where weights for each output channel share the same quantization range.

Table \ref{table:quantization summary} summarizes the different quantization schemes for CPU and accelerator. 

\begin{table}[h!]
\centering
\begin{tabular}{l|rr}
 & CPU & Accelerator \\
\hline
Weights \& Emb & INT4 sym, group-size 32 & INT4 sym, channel-wise \\
Activations & INT8 dyn asym per-token & BF16 \\
KV cache & INT8 dyn asym per-token & BF16 \\
QAT algorithm & Vanilla & Learnable quant ranges \\
\end{tabular}
\caption{Comparison of quantization schemes for CPU and accelerator. Emb=Embeddings, dyn=dynamic, sym=symmetric}
\label{table:quantization summary}
\end{table}

\subsection{Quantization-Aware Training}
We found that post-training quantization (PTQ) causes a severe quality regression when directly applied to the Phase 3 pre-training checkpoint (see Table \ref{table:PTQ}). Therefore, we utilize quantization-aware training (QAT) to more effectively quantize our pre-trained model for on-device use-cases \citep{krishnamoorthi2018quantizing,nagel2021white,liu2023llm}. Here, the full precision checkpoint is trained to be resilient to quantization by keeping quantization operations in-the-loop. In our QAT setup, all linear layer weights are quantized to INT4, symmetrically within the range of $[-8, 7]$ and subsequently de-quantized to enable backpropagation through the quantization operation described below:
\begin{align}
    Q(W_g) &= s_w \times \text{clip} \left(\text{round}\left(\frac{W_g}{s_{w}} \right),-8,7 \right),  \quad s_w = \frac{1}{7.5} \text{max} \left( -w_{min}, w_{max} \right)
\end{align}
where $W_g$ refers to a group of weights for a particular output neuron and $s_w$ is the corresponding scaling factor. There are multiple alternatives on how the $w_{min}$ and $w_{max}$ bounds are treated. Here, we explore two flavors, defined as:

\begin{enumerate}
\item \textbf{Vanilla:} The ranges $w_{win}$ and $w_{max}$ are computed from $W_g$: $w_{max} = \max W_g, w_{min} = \min W_g$
\item \textbf{Learnable:} The ranges $w_{min}$ and $w_{max}$ are learned as part of the optimization process with standard backpropagation
\end{enumerate}

 Since channel-wise quantization is more coarse-grained compared to group-wise quantization, we find that learnable range QAT is necessary to close the quality gap between the channel-wise quantized and floating-point models (see Table \ref{table:range learning vs dynamic})

By quantizing the weights and embeddings to INT4, we are able to reduce the model size from 2.2 GB (using bf16 weights in the full precision model) to 590 MB for the CPU model. Since accelerators do not typically support weight sharing, we do not share weights between the embedding and unembedding layers, making the accelerator model 720 MB.

In line with previous pre-training phases, the QAT phase also incorporates long-context awareness, however, instead of the Llama 4-Scout teacher model previously used for knowledge distillation, we use self-knowledge distillation in the QAT stage with the full-precision Phase 3 model as the teacher model (see Table \ref{table:self KD}). Furthermore, as typically done in the final QAT training stage, we limit the number of steps and tokens of this training step to a small fraction of the full-precision training. Specifically, we performed QAT for approximately 5\% of of the full-precision training budget, utilizing only 80 billion tokens.

The quantized checkpoint are available at \url{https://huggingface.co/facebook/MobileLLM-Pro-base-int4-cpu} and \url{https://huggingface.co/facebook/MobileLLM-Pro-base-int4-accelerator}

\section{Pre-Training Results}
\subsection{Main Results}
\label{sec:pt_main_results}
We evaluate our pre-trained checkpoint across reasoning, long-context retrieval, and knowledge-intensive tasks. Specifically, we assess the performance along the same dimensions as our two main baselines: Gemma 3-1B~\citep{gemmateam2025gemma3technicalreport} and Llama 3.2-1B~\citep{touvron2024llama3} in Table \ref{tab:main_results_pt}.

\begin{table}[h!]
\centering
\begin{tabular}{l|l|rrr}
\textbf{Benchmark}      & \textbf{Metric} & \textbf{MobileLLM-Pro} & \textbf{Gemma 3-1B} & \textbf{Llama 3.2-1B} \\
\hline
HellaSwag \citeyearpar{zellers-etal-2019-hellaswag}    & acc\_char & \textbf{67.11} & 62.30 & 65.69 \\
BoolQ   \citeyearpar{clark-etal-2019-boolq}      & acc\_char & \textbf{76.24} & 63.20 & 62.51 \\
PIQA   \citeyearpar{bisk2020piqa}       & acc\_char & \textbf{76.55} & 73.80 & 75.14 \\
SIQA   \citeyearpar{sap2019socialiqa}       & acc\_char & \textbf{50.87} & 48.90 & 45.60 \\
TriviaQA   \citeyearpar{joshi2017triviaqa}     & em        & \textbf{39.85} & 39.80 & 23.81 \\
Natural Questions \citeyearpar{kwiatkowski2019natural}& em        & \textbf{15.76} & 9.48  & 5.48  \\
ARC-Challenge   \citeyearpar{clark2018arc}     & acc\_char & \textbf{52.62} & 38.40 & 38.28 \\
ARC-Easy   \citeyearpar{clark2018arc}      & acc\_char & \textbf{76.28} & 73.00 & 63.47 \\
WinoGrande  \citeyearpar{sakaguchi2021winogrande}  & acc\_char & \textbf{62.83} & 58.20 & 61.09 \\
OBQA    \citeyearpar{mihaylov2018openbookqa}      & acc\_char & \textbf{43.60} &     --  & 37.20 \\
NIH           & em        & \textbf{100.00} &    --  & 96.80 \\
\end{tabular}
\caption{Comparison of \methodname~against Gemma 3-1B and Llama 3.2-1B on eleven commonly used pre-training benchmarks at the 1B parameter scale (Best Value Per Row \textbf{Bold}, --=Not published).}
\label{tab:main_results_pt}
\end{table}

The results presented in Table~\ref{tab:main_results_pt} demonstrate that our model consistently outperforms both Gemma 3-1B and Llama 3.2-1B across a broad range of language understanding, long-context, and reasoning benchmarks. Notably, our model outperforms both baseline models on the average performance across tasks, as well as each individual metric. On many of the benchmarks, such as BoolQ, Natural Questions, and ARC-Challenge, \methodname~shows substantial improvements of over 5\% compared to both baselines. These results clearly highlight the effectiveness of our pre-training strategies and data curation efforts.
Regarding long-context tasks, utilizing implicit positional distillation, we achieve strong performance on the Needle In Haystack (NIH) benchmark, showing 100\% retrieval accuracy, surpassing Llama 3.2-1B by more than 3 percentage points. While performance on some datasets like TriviaQA only shows minor gains compared to Gemma 3-1B, we clearly achieve new state-of-the-art performance, establishing \methodname~as the most capable generalist langauge model in the 1B parameter class. 

\begin{table}[h!]
\centering
\begin{tabular}{l|rrr}
\textbf{Benchmark} & \textbf{Full Precision} & \textbf{Quant-CPU} & \textbf{Quant-Accelerator} \\
\hline
HellaSwag    & 67.11 & 64.89 & 65.10 \\
BoolQ        & 76.24 & 77.49 & 76.36 \\
PIQA         & 76.55 & 76.66 & 75.52 \\
SIQA    & 50.87 & 51.18 & 50.05 \\
TriviaQA       & 39.85 & 37.26 & 36.42 \\
Natural Questions         & 15.76 & 15.43 & 13.19 \\
ARC-Challenge        & 52.62 & 52.45 & 51.24 \\
ARC-Easy        & 76.28 & 76.58 & 75.73 \\
WinoGrande   & 62.83 & 62.43 & 61.96 \\
OBQA         & 43.60 & 44.20 & 40.40 \\
NIH          & 100.00 & 96.44 & 98.67 \\
\hline
\textbf{Average} & \textbf{61.81} & \textbf{61.08} & \textbf{60.42}
\end{tabular}
\caption{Comparison of our full-precision \methodname~results against CPU and accelerator quantized versions, showing minor regressions on the average pre-training performance.}
\label{table:full quantization results}
\end{table}

\subsection{Quantization Results}
Table~\ref{table:full quantization results} presents our full-precision model performance (discussed in section \ref{sec:pt_main_results}) against quantized versions optimized for CPU and accelerator hardware. Overall, the results indicate that quantization introduces only a minor performance reduction. Specifically, the average score drops from 61.81 using full precision to 61.08 for Quant-CPU and 60.42 for Quant-Accelerator, demonstrating that quantized models only regress $0.73\%$ and $1.3\%$ (absolute) compared to the full precision performance.

\begin{figure}[h!]
    \centering
    \includegraphics[width=0.7\linewidth]{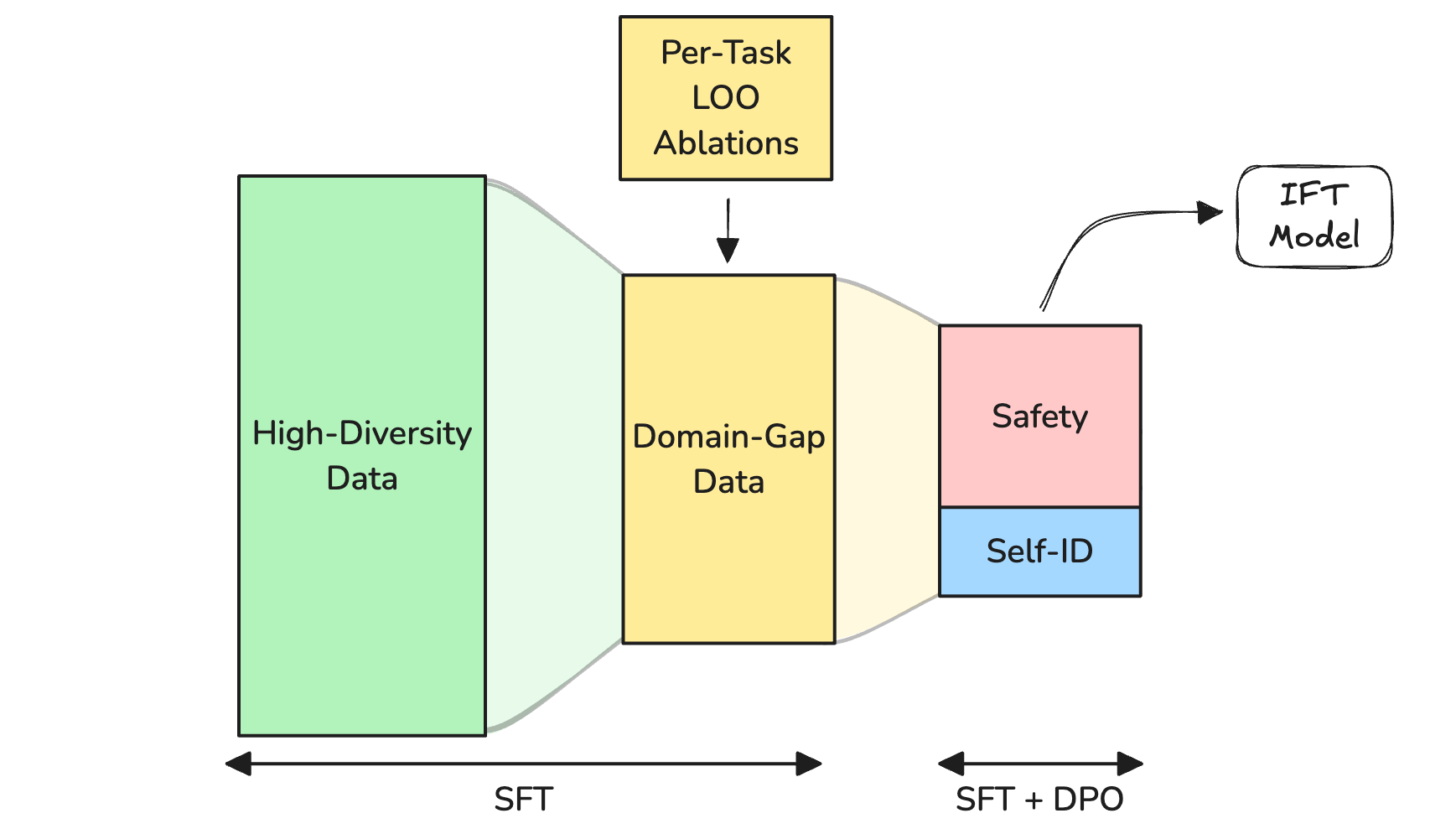}
    \caption{Overview of our three instruction fine-tuning phases. Similar to pre-training, the instruction training token budget progressively decreases across phases (indicated by the size of blocks). In the second instruction-tuning stage, we use external leave-one-out (LOO) information to better adjust the training data-mix.}
    \label{fig:ift_process}
\end{figure}

\section{Instruction Fine-Tuning}
\label{section:ift}
On top of our pre-trained \methodname\texttt{-base}~checkpoint, we showcase the model's capability as a general-purpose on-device assistant by providing an instruction fine-tuned checkpoint. We define the term ``assistant'' as a chat-based model that can support common on-device use-cases, such as general question-answering, function calling, in-context information retrieval, summarization, message rewriting, logical reasoning, and instruction following. 

For the instruction fine-tuned checkpoint, we continue to use the extended context length of $128,000$ tokens, and the efficient local-global self-attention, however, change the training objective to the standard cross entropy loss. We split the instruction fine-tuning process of our model into three distinct phases, as depicted in Figure \ref{fig:ift_process} and elaborated in the following sections.

\begin{table}[h!]
\centering
\begin{tabular}{l|r}
\textbf{Dataset}    & \textbf{Samples (M)} \\
\hline
Nemotron Math \citeyearpar{bercovich2025llamanemotronefficientreasoningmodel}          & 2.70 \\
Nemotron Safety  \citeyearpar{bercovich2025llamanemotronefficientreasoningmodel}       & 0.03 \\
Nemotron Code  \citeyearpar{bercovich2025llamanemotronefficientreasoningmodel}         & 0.65 \\
Nemotron Chat  \citeyearpar{bercovich2025llamanemotronefficientreasoningmodel}         & 0.04 \\
Nemotron Science  \citeyearpar{bercovich2025llamanemotronefficientreasoningmodel}      & 0.48 \\
Flan \citeyearpar{weifinetuned}               & 2.80 \\
Tulu 3    \citeyearpar{lambert2024tulu3}          & 0.94 \\
\hline
\textbf{Total} & \textbf{7.64}\\
\end{tabular}
\caption{Number of samples (M=millions) for each instruction fine-tuning dataset.}
\label{tab:ift_data}
\end{table}

\subsection{Diversity-first Instruction-Tuning}
In the first stage of instruction-tuning, we find in preliminary experiments that data diversity has a strong correlation with overall model performance. As a result, we utilize the natural distribution of our instruction fine-tuning data as guidance for the first stage data-mix, sticking closely to the sample-count of our open-source datasets shown in Table \ref{tab:ift_data}. Given the large discrepancy in sample count between domains, the model is mostly exposed to large amounts of Nemotron Math and Flan datapoints, accounting for around 70\% of overall samples in this phase. Despite this imbalance in datasets and domains, we find that in this initial phase, oversampling smaller datasets has a negative effect on model performance. 

\subsection{Leave-One-Out Continued Tuning}
\begin{figure}[h!]
    \centering
    \includegraphics[width=0.75\linewidth]{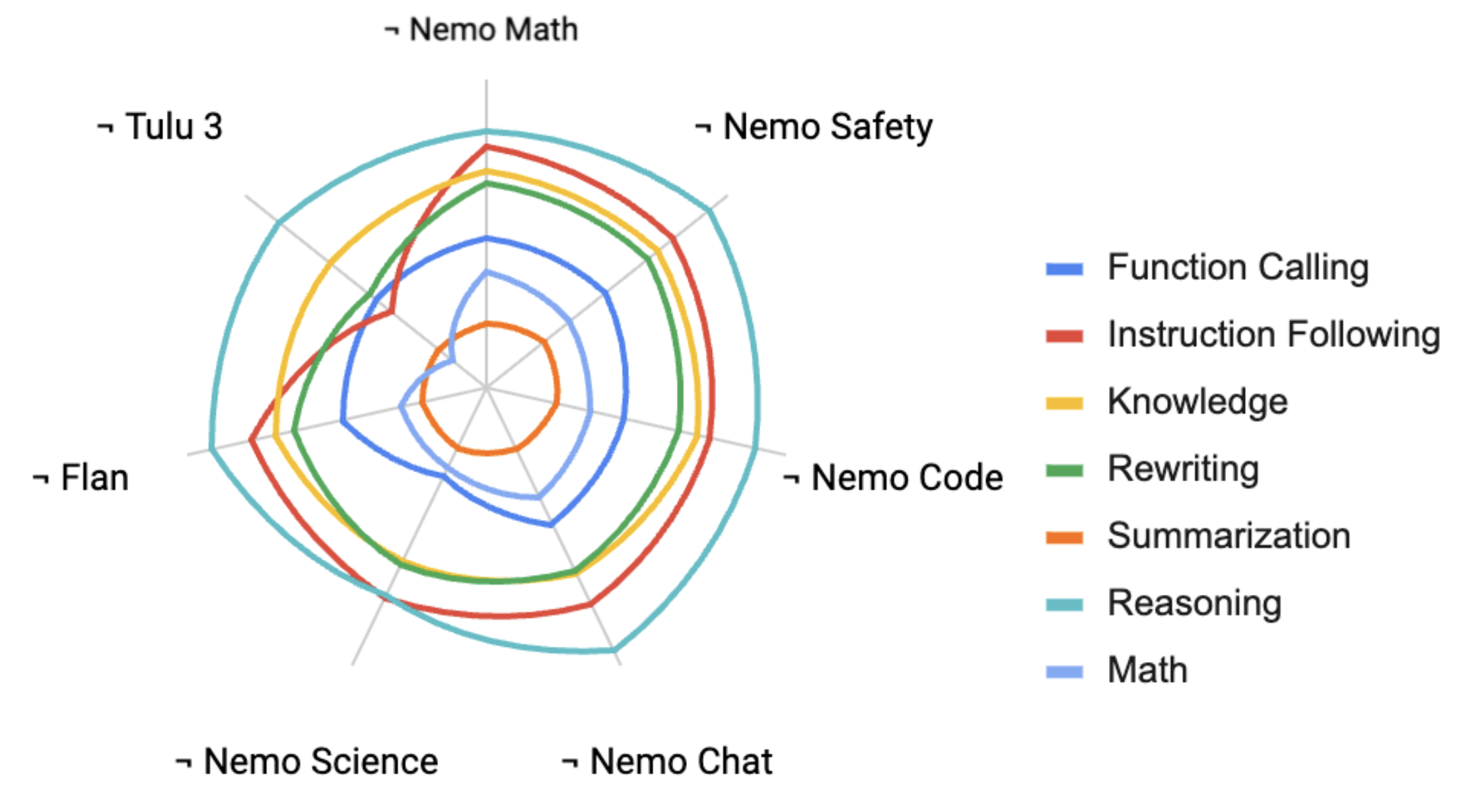}
    \caption{Radar Chart showing the result of our leave-one-out (LOO) ablations in the second instruction fine-tuning (IFT) stage. Each dimension describes the performance with the respective dataset held out (e.g. $\neg$Tulu 3 indicates the ablation performance of all IFT datasets, except Tulu 3) across domain-specific abilities (see Legend).}
    \label{fig:radar}
\end{figure}

In the second phase of instruction fine-tuning, we tailor the data-mix to the final Phase 1 model checkpoint. As such, we run a leave-one-out (LOO) experiment on the 7 IFT domains in Table \ref{tab:ift_data} to quantify the impact of individual domains on our desired abilities. Figure \ref{fig:radar} shows the leave-one-out results, indicating that the Tulu 3 and Nemotron Science data-patches have the largest impact on model regressions. In accordance with this finding, we adjust the second stage data-mix to further improve the overall model performance.

\subsection{Safety and Self-Identification Alignment}
In the final instruction fine-tuning stage, we aim to improve model safety and self-identification. To this end, we combine supervised fine-tuning (SFT) and direct preference optimization (DPO, \citet{rafailov2024directpreferenceoptimizationlanguage}) methodologies to anneal the model using synthetic data. To achieve the desired level of model safety and self-identification, we trade-off benchmark performance, resulting in our final instruction-tuned checkpoint.

Our instruction-tuned checkpoint is available at: \url{https://huggingface.co/facebook/MobileLLM-Pro}

\section{Instruction Fine-Tuning Results}
\subsection{Main Results}
\label{sec:ift_main_results}
Our instruction fine-tuning evaluation is conducted across nine chat-based benchmarks, which can be broadly categorized into seven categories. Namely, we evaluate the model's knowledge (MMLU), instruction following (IFEval), coding (MBPP, HumanEval), question-answering (ARC-Challenge, HellaSwag), Function-Calling (BFCLv2), rewriting (OpenRewrite) and summarization (TLDR9+) abilities. Each task is formatted as a user-assistant style conversation and evaluated on the final model response. The experimental results presented in Table \ref{tab:ift_results} demonstrate that \methodname~achieves overall strong instruction-tuning performance across this diverse set of evaluations. Specifically, our model outperforms Gemma 3-1B and Llama 3.2-1B on MBPP, HumanEval, ARC-Challenge, HellaSwag, BFCL v2, Open Rewrite, and TLDR9+, while beating one of the two baselines on both, MMLU and IFEval. Despite not reaching superior performance across all benchmarks, \methodname~shows promising performance across meaningful ``assistant-style'' dimensions, making it the best available 1B instruction-tuned model for a wide range of on-device assistant tasks.

\begin{table}[h!]
\centering
\begin{tabular}{l|l|rrr}
\textbf{Benchmark} & \textbf{Metric} & \textbf{MobileLLM-Pro} & \textbf{Gemma 3-1B} & \textbf{Llama 3.2-1B} \\
\hline
MMLU    \citeyearpar{hendrycks2021measuringmassivemultitasklanguage}     & macro\_avg/acc & \underline{44.8}  & 29.9  & \textbf{49.3} \\
IFEval   \citeyearpar{zhou2023instructionfollowingevaluationlargelanguage}    & acc* & \underline{62.0}  & \textbf{80.2} & 59.5 \\
MBPP    \citeyearpar{austin2021programsynthesislargelanguage}     & pass@1 & \textbf{46.8} & 35.2  & \underline{39.6} \\
HumanEval  \citeyearpar{chen2021evaluatinglargelanguagemodels}  & pass@1 & \textbf{59.8} & \underline{41.5}  & 37.8 \\
ARC-Challenge   \citeyearpar{clark2018arc}     & acc & \textbf{62.7} &--& 59.4     \\
HellaSwag  \citeyearpar{zellers-etal-2019-hellaswag}  & acc & \textbf{58.4} &--& 41.2    \\
BFCL v2   \citeyearpar{patil2025bfcl}   & acc & \textbf{29.4} &--& 25.7    \\
Open Rewrite \citeyearpar{shu2023rewritelminstructiontunedlargelanguage} & micro\_avg/rougeL & \textbf{51.0} &--& 41.6   \\
TLDR9+  \citeyearpar{sotudeh-etal-2021-tldr9}     & rougeL & \textbf{16.8} &--& \textbf{16.8} \\
\end{tabular}
\caption{Instruction tuning benchmark results for \methodname~(Instruct), Gemma 3-1B (Instruct), and Llama 3.2-1B (Instruct). \textbf{Bold} indicates best result, \underline{underline} shows second best model for benchmarks with more than two comparisons. *=Average accuracy on Prompt/Instruction set for Strict/Loose.}
\label{tab:ift_results}
\end{table}

\subsection{Human Evaluation}
While public benchmarks are an important indicator for general model abilities, they oftentimes fall short in truly describing the potential of foundational language models, due to potential data contamination, hill-climbing, and limited generalization based on a small evaluation sets. To supplement our benchmark results presented in section \ref{sec:ift_main_results}, we provide additional human evaluation results directly comparing model responses between model candidates.

Human evaluations are conducted by feeding 100 sample prompts for each of the four benchmark dimensions (Summarization, Rewriting, Recall, and Tool Calling) into the competing models and have human raters give a ternary preference label between them. Each prompt is single-annotated with one of three classes identifying ``Model A wins'', ``Model B wins'' or ``Tie''. The order of models presented to the human annotators are shuffled on instance level. The results of the annotations are shown in Tables \ref{tab:HE_Gemma} and \ref{tab:HE_Llama}, comparing \methodname~against Gemma 3-1B and Llama 3.2-1B respectively\footnote{Please note that due to the strict form of tool calling responses and the resulting verifiability, we replace the human rater with Llama 3-70B as the judge for this task.}. 

\begin{table}[h!]
\centering
\begin{tabular}{l|rrr}
\textbf{Task} & \textbf{MobileLLM-Pro} & \textbf{Gemma 3-1B} & \textbf{Tie} \\
\hline
Summarization & 47 & \textbf{51} & 2  \\
Rewrite       & 45 & \textbf{49} & 6 \\
Recall        & \textbf{47} & 30 & 23  \\
Tool Calling* & \textbf{53} & 24	& 23 \\
\end{tabular}
\caption{Human Evaluation of \methodname~against Gemma 3-1B on 100 samples per category. \textbf{Bold} shows winner. * = Using LLM evaluation instead of humans due to verifiability}
\label{tab:HE_Gemma}
\end{table}

\begin{table}[h!]
\centering
\begin{tabular}{l|rrr}
\textbf{Task} & \textbf{MobileLLM-Pro} & \textbf{Llama 3.2-1B} & \textbf{Tie} \\
\hline
Summarization & \textbf{42} & 30 & 28  \\
Rewrite       & \textbf{53} & 30 & 16  \\
Recall        & \textbf{56} & 8  & 37  \\
Tool Calling* & \textbf{40} & 35 & 25 \\

\end{tabular}
\caption{Human Evaluation of \methodname~against Llama 3.2-1B on 100 samples per category. \textbf{Bold} shows winner. * = Using LLM evaluation instead of humans due to verifiability}
\label{tab:HE_Llama}
\end{table}

Looking at the results we find that the human evaluations corroborate our findings in public benchmarks. Specifically, \methodname~shows large improvements over Llama 3.2-1B across all evaluation dimensions, with a particularly strong lead in the recall and tool calling comparison.
Against Gemma 3-1B, \methodname~is preferred in the recall and tool calling dimension, while Gemma 3-1B wins in summarization and rewrite, however at a significantly smaller margin.

In summary, the human evaluation results collectively corroborate our benchmark results and highlight \methodname’s balanced strengths across assistant-style tasks.

\section{Latency Benchmarks}
\begin{table}[h!]
\centering
\begin{tabular}{l|c|rrr}
\textbf{Benchmark} & \textbf{Metric} & \textbf{2k} & \textbf{4k} & \textbf{8k} \\
\hline
CPU Prefill Latency & seconds & 8.9   & 24.8  & 63.5  \\
HTP Prefill Latency & seconds & 2.0  & 3.4  & 9.8  \\
CPU Decode Speed & toks/s & 33.6  & 24.8  & 19.7  \\
HTP Decode Speed & toks/s  & 31.6 & 29.0 & 22.8 \\
KV Cache Size & MB  & 14.0    & 23.0    & 40.0    \\
\end{tabular}
\caption{Prefill latency, decoding speed, and KV cache size for different input prompt lengths of 2k, 4k and 8k, along with decode speed of generating 1k tokens.}
\label{tab:latency-benchmarks}
\end{table}

To comprehensively evaluate \methodname's inference performance, we present a set of latency benchmarks in Table \ref{tab:latency-benchmarks} conducted on the Samsung Galaxy S25 CPU and the Samsung Galaxy S24 Hexagon Tensor Processor (HTP). For deployment, models were exported using ExecuTorch, leveraging the xnnpack backend for CPU execution and the HTP backend for accelerator runs. We show a set of three measures: Model prefill latency, measuring the time required to process the input prompt before generation begins, Decode speed, defined as the number of tokens generated per second during inference, and the KV cache size for different prompt lengths (2k, 4k, and 8k tokens). For each prompt length, the prefill latency and decode speed is presented for both, CPU and HTP.

The results in Table~\ref{tab:latency-benchmarks} demonstrate two trends: (1) As expected, both prefill latency and decode speed are dependent on the input prompt length and hardware choice. (2) HTP processing is consistently faster than CPU in both, prefill and decode speed, however, achieves significantly larger speed improvements during prefill.

\section{Ablations}
In this section, we show additional ablations around the main themes of this report. Specifically, we dive deeper into the impact of our data-mix, the decision to utilize knowledge distillation across pre-training stages, additional experiments to evaluate the benefit of implicit positional distillation, the impact of parallel expert checkpoint averaging in the final pre-training stage, and ablations around QAT.

\subsection{Data-mixing via Offline Simulation}
To better understand the impact of our offline simulation runs and the resulting data-mix, we ablate across multiple initial pre-training and instruction-tuning runs. To save compute, we run our data-mix ablations using regular cross-entropy training in both, the pre-training and instruction-tuning setup, instead of using the more computationally heavy knowledge distillation approach. The comparison between the uniform data-mix and our Scalable Data Mixer (SDM) \citep{chang2025sdm} weights is shown in Table~\ref{tab:data-mix_comparison}. 

\begin{table}[h!]
\centering
\vspace{0.5em}
\resizebox{0.7\textwidth}{!}{%
\begin{tabular}{lccc}
\toprule
\textbf{Training Stage} & \textbf{Dataset / Tokens} & \textbf{data-mix Source} & \textbf{Avg. Performance (\%)} \\
\midrule
pre-training 
  & 1.4T & uniform & 38.70 \\
  &  & SDM & \textbf{49.31} \\
\midrule
Instruction Tuning 
  & 80B & uniform & 17.94 \\
  &  & SDM & \textbf{45.23} \\
\bottomrule
\end{tabular}%
}
\caption{Comparison of model performance between uniform data-mixing and our SDM data-mixing strategy. 
Results show consistent gains across pre-training and instruction tuning stages. Pre-training results shown are obtained using cross-entropy loss. 
SDM denotes the optimized data-mix with reweighted sources of OSS data.}
\label{tab:data-mix_comparison}
\end{table}

The average performance comparison between the two settings for both, pre-training and instruction-tuning tells a clear story: Utilizing enhanced data-mixes results in substantial improvements throughout training. During pre-training, performance increased over 10\% absolute, from 38.7\% to 49.3\%, highlighting the effectiveness of adaptive weighting in selecting linguistically diverse yet relevant data sources. Even more impressive, in the instruction-tuning stage, where SDM points towards maximizing data diversity in the initial phase, we are able to boost the average benchmark performance by over 15\%, from 17.9\% to 32.7\%, suggesting that the learned data weights better aligns with the downstream objectives. Overall, these results indicate that an offline, utility-driven data-mixing approach yields consistent quality gains across stages of model development.

\subsection{Training Objectives -- The Impact of Knowledge Distillation during Pre-Training} To assess the impact of knowledge distillation on the model's pre-training performance, we compare a model instance trained using cross-entropy loss against a candidate using logit-based knowledge distillation from Llama 4-Scout. For a fair comparison between the two approaches, we FLOP-align the training process, reducing the number of training steps for the KD model candidate to account for the increased FLOP count attributed to the teacher model forward pass. In the evaluation shown in Table \ref{tab:abl_ce_kd}, we find that knowledge distillation is clearly improving our Stage 1 pre-training performance, leading to an average performance improvement of 4.4\% absolute compared to cross-entropy pre-training.

\begin{table}[h!]
\centering
\begin{tabular}{l|r}
\textbf{Approach} & \textbf{Avg. Performance} \\
\hline
CE & 49.31 \\
KD & 53.74 \\
\end{tabular}
\caption{Average Phase 1 ablation performance of cross entropy and knowledge distillation setups. We exclude the NIH metric from this comparison, since the model has not been trained on long context sequences.}
\label{tab:abl_ce_kd}
\end{table}

\subsection{Implicit Positional Distillation} As one of the main novelties of our work, we replace long-context datasets, commonly used to extend the model's effective context support, by implicit positional distillation. To better quantify the impact of the implicit positional distillation method, we compare the final Stage 1 checkpoint (before long-context extension) with an ablation utilizing distinct long context data, and our implicit positional distillation. Table \ref{tab:long_cxt_abl} shows the resulting benchmark performance of the three experiments. We find that the final Phase-1 checkpoint (trained with $2048$ context length) does lack long-context ability. For the Phase 2 comparison, we find that using dedicated long-context data significantly improves the long-context performance compared to the final Phase 1 checkpoint, reaching 80.22\% on the NIH metric, however, does significantly regress in other benchmark dimensions (-5.9\%). Looking at our  implicit positional distillation experiment, we see that utilizing the same pre-training data as in Phase 1, hence, removing the data distribution shift between phases, the model achieves high scores on both, NIH and the average pre-training benchmarks, regressing less than half a percent point.

\begin{table}[h!]
\centering
\begin{tabular}{l|rr}
\textbf{Stage} & \textbf{NIH Score} & \textbf{Avg. Performance} \\
\hline
Phase-1 & 6.7 & 53.74 \\
Phase-2 long context data & 80.22 & 47.86 \\
Phase-2 Implicit Pos Distillation & 99.78 & 53.57 \\
\end{tabular}
\caption{Phase-2 long-context methods and their impact on the Needle in Haystack (NIH) benchmark and average benchmark performance (excluding NIH) using distinct long context data.}
\label{tab:long_cxt_abl}
\end{table}

\subsection{Specialist Model Merging}
In the third pre-training phase, we anneal the model using our novel specialist model merging approach. As described in section \ref{sec:ph3}, we train multiple, parallel specialist branches on distinct data patches for a few hundred steps using a small learning rate, to not steer away too far from the initial model weights. In Table \ref{tab:spec_branch}, we show the pre-annealing, per-specialist and weight-averaged results of our model. Interestingly, while each specialist does improve the benchmark performance on average, individual specialists regress on some metrics. Surprisingly, the final weight averaged checkpoint consistently improves performance beyond the pre-annealed checkpoint and in most cases even outperforms the best specialist model. We believe that this phenomenon is caused by symbiotic relationships between specialists, which results in improved performance of the averaged model.

\begin{table}[h!]
\centering
\begin{tabular}{l|rrrrrr}
\textbf{Model} & \textbf{Pre-Anneal} & \textbf{Specialist 1}& \textbf{Specialist  2}& \textbf{Specialist 3}& \textbf{Specialist 4}& \textbf{Weight Avg.} \\
\hline
HellaSwag   & 65.77   & 66.69   & 66.87   & 66.31   & 67.04   & 67.58   \\
BoolQ       & 71.28   & 76.51   & 76.94   & 73.64   & 78.47   & 76.82   \\
PIQA        & 75.57   & 75.14   & 76.50   & 75.46   & 76.17   & 76.55   \\
SIQA        & 47.29   & 47.90   & 50.97   & 46.88   & 50.31   & 51.23   \\
TriviaQA    & 36.61   & 36.78   & 40.07   & 37.48   & 39.91   & 40.18   \\
Natural Questions & 12.02   & 12.22   & 15.84   & 11.83   & 16.01   & 16.29   \\
ARC-Challenge & 50.64  & 53.05   & 52.19   & 51.33   & 53.05   & 54.16   \\
ARC-Easy   & 71.37  & 75.22   & 76.45   & 73.66   & 76.74   & 76.96   \\
WinoGrande & 63.30  & 63.06   & 63.14   & 62.90   & 63.06   & 63.54   \\
OBQA       & 41.80  & 41.40   & 43.40   & 41.80   & 43.40   & 44.00   \\
NIH        & 100.00 & 100.00  & 100.00  & 100.00  & 100.00  & 100.00  \\
\hline
\textbf{Average} & \textbf{53.57} & \textbf{54.80} & \textbf{56.24} & \textbf{54.13} & \textbf{56.42} & \textbf{56.73} \\
\end{tabular}
\caption{Model performance before annealing compared to individual specialists and the final, model weight merged checkpoint. Average computed without NIH.}
\label{tab:spec_branch}
\end{table}

\subsection{Quantization Aware Training (QAT)}
We first ablate the impact of learning the quantization range for channel-wise quantization in Table \ref{table:range learning vs dynamic}. The results show that learnable ranges boost the average benchmark score by $4.79\%$ (absolute). 
\begin{table}[h!]
\centering
\begin{tabular}{l|rr}
\textbf{QAT algorithm} & \textbf{Avg. score} \\
\hline
Compute $w_{min/max}$    & 55.67 \\
Learn $w_{min/max}$    & 60.46 \\
\end{tabular}
\caption{Comparison of range-learning with computing $w_{min}, w_{max}$ from $W$ for channel-wise quantization.}
\label{table:range learning vs dynamic}
\end{table}

Table \ref{table:self KD} shows the benefit of knowledge distillation for QAT, bringing a significant $2.43\%$ (absolute) boost in average benchmark scores.
\begin{table}[h!]
\centering
\begin{tabular}{l|rr}
\textbf{KD setting} & \textbf{Avg. score} \\
\hline
No KD    & 58.61\\
KD from FP ckpt & 61.04\\
\end{tabular}
\caption{Comparison of QAT quality for the CPU model with and without self-knowledge-distillation from the full-precision (FP) teacher checkpoint.}
\label{table:self KD}
\end{table}

Finally, we quantified the benefit of QAT by comparing the CPU quantized model to a standard round-to-nearest post-training quantization (PTQ) baseline (Table \ref{table:PTQ}). The results show that QAT brings significant benefits in model quality.
\begin{table}[h!]
\centering
\begin{tabular}{l|rr}
\textbf{Quant setting} & \textbf{Regression relative to FP}\\
\hline
PTQ (round-to-nearest)    & 17\\
QAT & 0.4
\end{tabular}
\caption{Comparison of PTQ and QAT regression (absolute) in average benchmark scores relative to the unquantized (FP) model for the CPU model.}
\label{table:PTQ}
\end{table}

\section{Related Work}
\label{section:rel_work}
In recent years, the development of small foundational language models (under 3B parameters) has accelerated, motivated by the demand for efficient, deployable models that retain strong generalist capabilities. In recent years, efforts concentrated on scaling down transformer models through distillation from larger checkpoints, showing parameter savings without sacrificing performance, as seen in the LLaMA series \citep{touvron2023llama, touvron2024llama3} and the Gemma models \citep{anil2024gemma, gemmateam2025gemma3technicalreport}. These works showed that thoughtful architecture design and curated data can produce compact models competitive with much larger ones across diverse benchmarks. Similarly, Qwen 3 \citep{yang2025qwen3technicalreport} and SmolLM \citep{allal2025smollm2smolgoesbig} have introduced improvements at on-device scales along the training efficiency, data quality, and data mixing strategies.

A key challenge for small models is robust training and generalization, particularly in the face of data distribution shifts and limited capacity. Recent studies have highlighted the importance of high-quality, diverse pre-training data and advanced data mixing strategies. The Automixer framework \citep{chang2025automixercheckpointartifactsautomatic} introduced dynamic data weighting and mixing, enabling models to adapt to evolving data distributions and task requirements. Similarly, several small model pipelines, including SmolLM2 \citep{allal2025smollm2smolgoesbig} demonstrate that high quality source data and effective sampling strategies are critical for achieving strong generalization in compact models.

Long-context adaptation is another area of active research, with most small models historically limited to context windows below 32k tokens \citep{yang2025qwen3technicalreport, anil2024gemma, gemmateam2025gemma3technicalreport}. Recent work has explored architectural modifications and training techniques to extend context length without sacrificing efficiency. The Longformer \citep{beltagy2020longformer} and BigBird \citep{zaheer2020bigbird} architectures introduced sparse and global attention mechanisms, enabling efficient processing of long sequences. 

Finally, quantization and model compression have become essential for deploying small models on resource-constrained devices. Techniques such as GPTQ \citep{frantar2022gptq} and quantization-aware training (QAT) \citep{esser2019learned, jacob2018quantization} have enabled aggressive reduction in model size and inference latency, with minimal loss in accuracy. 

\section{Conclusion}
\label{section:conclusion}
In this work, we introduce \methodname, a 1-billion parameter foundational language model designed for efficient, high-quality on-device inference. To achieve this, our overall training pipeline contains a variety of novel techniques, especially tailored towards on-device size language model pre- and instruction-training. 

Namely, we utilize advanced data-mixing, implicit positional distillation, specialist model merging, and quantization-aware training to enable our on-device sized foundational language model to achieve state-of-the-art performance across a diverse set of pre-training benchmarks. In our evaluations, we show that \methodname~does not only surpass leading models like Gemma 3-1B and Llama 3.2-1B on standard pre-training benchmarks, but also delivers robust performance in instruction-following and assistant-style scenarios. Notably, to make \methodname~a true on-device model candidate, we show that our model maintains its competitive edge even after aggressive 4-bit quantization, enabling practical deployment on mobile CPUs and accelerators with minimal loss in accuracy.

Finally, releasing our methodologies, model checkpoints and code, we invite the research community to build on top of this line of research to further enhance on-device sized models. (\url{https://huggingface.co/collections/facebook/mobilellm-pro}).

\clearpage
\newpage
\bibliographystyle{assets/plainnat}
\bibliography{paper}

\end{document}